\documentclass[11pt,letterpaper]{article}
\usepackage{naaclhlt2016}
\usepackage{times}
\usepackage{latexsym}

\usepackage{graphicx}
\usepackage{amsmath}
\usepackage{url}
\usepackage{color, colortbl}
\usepackage{comment}

\definecolor{Gray}{gray}{0.9}
\newcolumntype{g}{>{\columncolor{Gray}}c}

\naaclfinalcopy % Uncomment this line for the final submission
\def\naaclpaperid{***} %  Enter the naacl Paper ID here

\title{Combining Neural Networks and Log-linear Models \\ to Improve Relation Extraction}

\author{Thien Huu Nguyen \\
  Computer Science Department \\
  New York University \\
  New York, NY 10003 USA \\
  {\tt thien@cs.nyu.edu} \\\And
  Ralph Grishman \\
  Computer Science Department \\
  New York University \\
  New York, NY 10003 USA \\
  {\tt grishman@cs.nyu.edu} \\}

\date{}

\begin{document}

\maketitle

\begin{abstract}

The last decade has witnessed the success of the traditional feature-based method on exploiting the discrete structures such as words or lexical patterns to extract relations from text. Recently, convolutional and recurrent neural networks has provided very effective mechanisms to capture the hidden structures within sentences via continuous representations, thereby significantly advancing the performance of relation extraction. The advantage of convolutional neural networks is their capacity to generalize the consecutive $k$-grams in the sentences while recurrent neural networks are effective to encode long ranges of sentence context. This paper proposes to combine the traditional feature-based method, the convolutional and recurrent neural networks to simultaneously benefit from their advantages. Our systematic evaluation of different network architectures and combination methods  demonstrates the effectiveness of this approach and results in the state-of-the-art performance on the ACE 2005 and SemEval dataset.

\end{abstract}

\section{Introduction}

We studies the relation extraction (RE) problem, one of the important problem of information extraction and natural language processing (NLP). Given two entity mentions in a sentence (relation mentions), we need to identify the semantic relationship (if any) between the two entity mentions. One example is the recognition of the {\it Located} relation between ``{\it He}'' and ``{\it Texas}'' in the sentence ``{\it He lives in Texas}''.

The two methods dominating RE research in the last decade are the feature-based method \cite{Kambhatla:04,Boschee:05,Zhou:05,Grishman:05,Jiang:07,Chan:10,Sun:11} and the kernel-based method \cite{Zelenko:03,Culotta:04,Bunescu:05a,Bunescu:05b,Zhang:06,Zhou:07,Qian:08,Nguyen:09,Plank:13}.  These research extensively studies the leverage of linguistic analysis and knowledge resources to construct the feature representations, involving the combination of {\it discrete} properties such as lexicon, syntax, gazetteers. Although these approaches are able to exploit the symbolic (discrete) structures within relation mentions, they also suffer from the difficulty to generalize over the unseen words \cite{Gormley:15}, motivating some very recent work on employing the {\it continuous} representations of words (word embeddings) to do RE. The most popular method involves neural networks (NNs) that effectively learn hidden structures of relation mentions from such word embeddings, thus achieving the top performance for RE \cite{Zeng:14,Santos:15,Xu:15}.

%ranging from the direct utilization of word embeddings as features in such traditional methods \cite{Plank:13,Nguyen:14,Nguyen:15b} to the initialization for the neural networks (NNs).

% The rationale of NNs for RE is to learn the hidden and effective structures of relation mentions from the initial word embeddings to improve the performance.

%This work follows the latter approach as it provides the mechanism to fine-tune the initial word embeddings to the specific task and better capture the hidden structures in the relation mentions \cite{Zeng:14,Santos:15,Xu:15}.

%Zeng:14,
%,Ebrahimi:15,Liu:15

The NN research for relation extraction and classification has centered around two main network architectures: convolutional neural networks (CNNs) \cite{Santos:15,Zeng:15} and recursive/recurrent neural networks \cite{Socher:12,Xu:15}. The distinction between convolutional neural networks and recurrent neural networks (RNNs) for RE is that the former aim to generalize the local and consecutive context (i.e, the $k$-grams) of the relation mentions \cite{Nguyen:15a} while the latter adaptively accumulate the context information in the whole sentence via the memory units, thereby encoding the global and possibly unconsecutive patterns for RE \cite{Hochreiter:97,Cho:14}. Consequently, the traditional feature-based method (i.e, the log-linear or MaxEnt model with hand-crafted features), the CNNs and the RNNs tend to focus on different angles for RE. Guided from this intuition, in this work, we propose to combine the three models to further improve the performance of RE.

% Besides, CNNs and RNNs differ from the traditional feature-based method (the log-linear model with hand-crafted features) in that CNNs and RNNs 

%This leads to the divergence in terms of patterns being captured for RE. 

While the architecture design of CNNs for RE is quite established due to the extensive studies in the last couple of years, the application of RNNs to RE is only very recent and the optimal designs of RNNs for RE are still an ongoing research. In this work, we first perform a systematic exploration of various network architectures to seek the best RNN model for RE. In the next step, we extensively study different methods to assemble the log-linear model, CNNs and RNNs for RE, leading to the combined models that yield the state-of-the-art performance on the ACE 2005 and SemEval dataset. To the best of our knowledge, this is the first work to systematically examine the RNN architectures as well as combine them with CNNs and the traditional feature-based approach for RE.

%Finally, we analyze the systems to better understand the operation of the models for RE.

%To the best of our knowledge, there is only one work from Liu et al. \shortcite{Liu:15} that 

%From the descriptions of CNNs and RNNs in Section \ref{sec:separateModel}, we see that the hidden vectors $\mathbf{s}$ in CNNs aims to generalize the local and consecutive context (i.e, the $k$-grams). The hidden vectors $h_1,h_2,\ldots,h_n$ in RNNs, in contrast, adaptively accumulate the context information in the whole sentence via the memory units, thereby encoding the global and possibly unconsecutive patterns for RE. This divergence in terms of patterns being captured makes it promising to combine CNNs and RNNs to further improve the performance of each model. In this work, we investigate three different methods to assemble CNNs and RNNs: ensembling, stacking and voting.

%The distinction between recursive and recurrent neural networks is that the former recur over the tree structures of sentences while the latter work directly on the sequential forms of the sentences. 

\section{Models}

%RE is formalized as a multi-class classification problem whose goal is to predict the semantic relationship between two entity mentions of relation mentions. 

Relation mentions consist of sentences marked with two entity mentions of interest. In this paper, we examine two different representations for the sentences in RE: (i) the standard representation, called SEQ that takes all the words in the sentences into account and (ii) the dependency representation, called DEP that only considers the words along the dependency paths between the two entity mention heads of the sentences. In the following, unless indicated specifically, all the statements about the sentences hold for both representations SEQ and DEP.

%We note that the previous studies on relation extraction/classification only assume either sentence representation \cite{Zeng:14,Mo:14,Gormley:15,Xu:15}. This is the first study comparing the two representations for sentence in the same setting.

Throughout this paper, for convenience, we assume that the input sentences of the relation mentions have the same fixed length $n$. This can be achieved by setting $n$ to the length of the longest input sentences and padding the shorter sentences with a special token. Let $W = w_1w_2\ldots w_n$ be the input sentence of some relation mention, where $w_i$ is the $i$-th word in the sentence. Also, let $w_{i_1}$ and $w_{i_2}$ be the two heads of the two entity mentions of interest. In order to prepare the relation mention for neural networks, we first transform each word $w_i$ into a real-valued vector $x_i$ using the concatenation of the following seven vectors, motivated by the previous research on neural networks and feature analysis for RE \cite{Zhou:05,Sun:11,Gormley:15}.

- The real-valued word embedding vector $e_i$ of $w_i$, obtained by  looking up the word embedding table $E$.
%: $e_i = E[w_i]$. 

%$E$ is often pre-trained by some language model on a large corpus to capture the latent syntactic and semantic properties of  words \cite{Bengio:03,Mnih:08,Turian:10}. 

- The real-valued distance embedding vectors $d_{i_1}$, $d_{i_2}$ to encode the relative distances $i-i_1$ and $i-i_2$ of $w_i$ to the two entity heads of interest $w_{i_1}$ and $w_{i_2}$: $d_{i_1} = D[i-i_1]$, $d_{i_2} = D[i-i_2]$ where $D$ is the distance embedding table (initialized randomly). The objective is to inform the networks the positions of the two entity mentions for relation prediction.

- The real-valued embedding vectors for entity types $t_i$ and chunks $q_i$ to embed the entity type and chunking information for $w_i$. These vectors are generated by looking up the entity type and chunk embedding tables (also initialized randomly) (i.e, $T$ and $Q$ respectively) for the entity type $ent_i$ and chunking label $chunk_i$ of $w_i$: $t_i = T[ent_i]$, $q_i = Q[chunk_i]$.

%, to be retrieved from human annotation or some automatic taggers

- The binary vector $p_i$ with one dimension to indicate whether the word $w_i$ is on the dependency path between $w_{i_1}$ and $w_{i_2}$ or not.

- The binary vector $g_i$ whose dimensions correspond to the possible relations between words in the dependency trees. The value at a dimension of $g_i$ is only set to 1 if there exists one edge of the corresponding relation connected to $w_i$ in the dependency tree.

%Note that the seven vectors above encompass different information of relation mentions (i.e, word embeddings, entity mention positions, entity types, chunking, dependency parsing) that have been shown to be useful for RE in the previous research \cite{Zhou:05,Sun:11,Nguyen:15,Gormley:15, blabla}.

%of dimension $d$

The transformation from the word $w_i$ to the vector $x_i = [e_i, d_{i_1}, d_{i_2}, t_i,q_i, p_i, g_i]$ essentially converts the relation mention with the input sentence $W$ into a real-valued matrix $X = [x_1,x_2,\ldots, x_n]$, to be used by the neural networks presented below.

% Note that the embedding tables $E, D, T, Q$ are also optimized during the training of the networks.

\subsection{The Separate Models}

\label{sec:separateModel}

We describe two typical NN architectures for RE underlying the combined models in this work.

\subsubsection{The Convolutional Neural Networks}

\label{sec:cnn}

In CNNs \cite{Kalchbrenner:14,Kim:14}, given a window size of $k$, we have a set of $c_k$ feature maps (filters). Each feature map $\mathbf{f}$ is a weight matrix $\mathbf{f} = [\mathbf{f}_1, \mathbf{f}_2, \ldots, \mathbf{f}_{k}]$ where $\mathbf{f}_i$ is a vector to be learnt during training as the model parameters. The core of CNNs is the application of the convolutional operator on the input matrix $X$ and the filter matrix $\mathbf{f}$ to produce a score sequence $\mathbf{s}^\mathbf{f} = [s^\mathbf{f}_1, s^\mathbf{f}_2, \ldots, s^\mathbf{f}_{n-k+1}]$, interpreted as a more abstract representation of the input matrix $X$:
$$s^\mathbf{f}_i = g(\sum_{j=0}^{k-1}\mathbf{f}_{j+1}x_{j+i} + b)$$

where $b$ is a bias term and $g$ is the $\tanh$ function.

% is some non-linear function (the $\tanh$ function in our case).

%In the next step, we further abstract the scores in $\mathbf{s}$ by aggregating it via the $\max$ function (max-pooling), generating a single number $\mathbf{s}_{\mathbf{f}}$ to represent the input sentence for the current feature map $\mathbf{f}$:
%$$\mathbf{s}^\mathbf{f}_{max} = \max\{s^\mathbf{f}_1, s^\mathbf{f}_2, \ldots, s^\mathbf{f}_{n-k+1}\}$$

In the next step, we further abstract the scores in $\mathbf{s}^\mathbf{f}$ by aggregating it via the $\max$ function to obtain the max-pooling score $\mathbf{s}^\mathbf{f}_{max}$. We then repeat this process for all the $c_k$ feature maps with different window sizes $k$ to generate a vector of the max-pooling scores. In the final step, we pass this vector into some standard multilayer neural network, followed by a softmax layer to produce the probabilistic distribution $p_{\text{C}}(y|X)$ over the possible relation classes $y$ in the prediction task.

%, generating a single number $\mathbf{s}^\mathbf{f}_{max}$ to represent the input sentence for the current feature map.
% $\mathbf{f}$
%$$\mathbf{s}^\mathbf{f}_{max} = \max\{s^\mathbf{f}_1, s^\mathbf{f}_2, \ldots, s^\mathbf{f}_{n-k+1}\}$$

%The process presented above can be then replicated for all the $c_k$ feature maps that output $c_k$ numbers to form a sentence representation vector $v_k$ of dimension $c_k$. We can also apply this process for different window sizes $k$, resulting in different representation vectors $v_k$ to cover various $k$-grams in the sentence \cite{Kalchbrenner:14}. Finally, the concatenation vector $v^C$ of the vectors $v_k$ for all the specified window sizes $k$ is utilized as the feature vector input in some standard multilayer neural network, followed by a softmax layer to produce the probabilistic distribution $p_{\text{C}}(y|X)$ over possible relation classes $y$ in our prediction task.

%The vector in the last layer of this multilayer neural network is called $z^C$ whose dimension is the number of relation classes (labels) $n_c$ in our prediction task. $z^C$ would be used later in our discussion of the combination models.

\subsubsection{The Recurrent Neural Networks}

\label{sec:rnn}

In RNNs, we consider the input matrix $X = [x_1,x_2,\ldots, x_n]$ as a sequence of column vectors indexed from 1 to $n$. At each step $i$, we compute the hidden vector $h_i$ from the current input vector $x_i$ and the previous hidden vector $h_{i-1}$ using the non-linear transformation function $\Phi$: $h_i = \Phi(x_i, h_{i-1})$.

This recurrent computation can be done via three different directional mechanisms: (i) the forward mechanism that recurs from 1 to $n$ and generate the forward hidden vector sequence: $R(x_1,x_2, \ldots, x_n) = h_1,h_2, \ldots, h_n$, (ii) the backward mechanism that runs RNNs from $n$ to 1 and results in the backward hidden vector sequence $R(x_n,x_{n-1}, \ldots, x_1) = h'_n,h'_{n-1}, \ldots, h'_1$\footnote{The initial hidden vectors are set to the zero vector.}, and (iii) the bidirectional mechanism that performs RNNs in both directions to produce the forward and backward hidden vector sequences, and then concatenate them at each position to generate the new hidden vector sequence $h^b_1,h^b_2, \ldots, h^b_n$: $h^b_i = [h_i, h'_i]$.

%feed $v^R$ into some standard multilayer neural network with a softmax layer at the end, resulting in the distribution $p_{\text{R}}(y|X)$ for the RNN models

Given the hidden vector sequence $h_1, h_2, \ldots, h_n$ obtained from one of the three mechanisms above, we study two following strategies to generate the representation vector $v^R$ for the initial relation mention. Note that this representation vector can be again fed into some standard multilayer neural network with a softmax layer in the end, resulting in the distribution $p_{\text{R}}(y|X)$ for the RNN models:

%\begin{itemize}
- The {\it HEAD} strategy: In this strategy, $v^R$ is the concatenation of the hidden vectors at the positions of the two entity mention heads of interest: $v^R = [h_{i_1}, h_{i_2}]$. This is motivated by the importance of the two mention heads in RE \cite{Sun:11,Nguyen:14}.

%Note that due to the recurrent mechanism, $h_{i_1}$ and $h_{i_2}$ not only have the information about the two entity heads but also contain the context information in the sentence.

- The {\it MAX} strategy: This strategy is similar to our max-pooling mechanism in CNNs. In particular, $v_R$ is obtained by taking the maximum along each dimension of the hidden vectors $h_1, h_2, \ldots, h_n$. The idea is to further abstract the hidden vectors by retaining only the most important feature in each dimension.

Regarding the non-linear function, the simplest form of $\Phi$ in the literature considers it as a one-layer feed-forward neural network, called $FF$: $h_i = FF(x_i,h_{i-1}) = \phi(Ux_i + Vh_{i-1})$ where $\phi$ is the {\it sigmoid} function. Unfortunately, the application of $FF$ causes the so-called ``{\it vanishing/exploding gradient}'' problems \cite{Bengio:94}, making it challenging to train RNNs properly \cite{Pascanu:12}. These problems are overcome by the long-short term memory units (LSTM) \cite{Hochreiter:97,Graves:09}.  In this work, we apply a variant of the memory units: the {\it Gated Recurrent Units}  from Cho et al. \shortcite{Cho:14}, called $GRU$. $GRU$ is shown to be much simpler than LSTM in terms of computation but still achieves the comparable performance \cite{Cho:14}.

\subsection{The Combined Models}

\label{sec:combinedModels}

%From the descriptions of CNNs and RNNs in Section \ref{sec:separateModel}, we see that the hidden vectors $\mathbf{s}$ in CNNs aims to generalize the local and consecutive context (i.e, the $k$-grams). The hidden vectors $h_1,h_2,\ldots,h_n$ in RNNs, in contrast, adaptively accumulate the context information in the whole sentence via the memory units, thereby encoding the global and possibly unconsecutive patterns for RE. This divergence in terms of patterns being captured makes it promising to combine CNNs and RNNs to further improve the performance of each model. 

We first present three different methods to assemble CNNs and RNNs: ensembling, stacking and voting, to be investigated in this work. The combination of the neural networks with the log-linear model would be discussed in the next section.

\subsubsection{Ensembling}

\label{sec:ensembling}

%(respectively) over the possible relation classes

In this method, we first run some CNN and RNN in Section \ref{sec:separateModel} over the input matrix $X$ to gather the corresponding distributions $p_{C}(y|X)$ and $p_{R}(y|X)$. We then combine the CNN and RNN by multiplying their distributions (element-wise): $p_{\text{ensemble}}(y|X) = \frac{1}{Z}p_{\text{C}}(y|X) p_{\text{R}}(y|X)$ ($Z$ is a normalization constant).

\subsubsection{Stacking}

The overall architecture of the stacking method is to use one of the two network architectures (i.e, CNNs and RNNs) to generalize the hidden vectors of the other architecture. The expectation is that we can learn more effective features for RE via such a deeper architecture by alternating between the local and global representations provided by CNNs and RNNs.

We examine two variants for this method. The first variant, called {\it RNN-CNN}, applies the CNN model in Section \ref{sec:cnn} on the hidden vector sequence generated by some RNN in Section \ref{sec:rnn} to perform RE. The second variant, called {\it CNN-RNN}, on the other hand, utilize the CNN model to acquire the hidden vector sequence, that is, in turn, fed as the input into some RNN for RE. For the second variant, as the length of the hidden vector  $\mathbf{s}^\mathbf{f} = [s^\mathbf{f}_1, s^\mathbf{f}_2, \ldots, s^\mathbf{f}_{n-k+1}]$ in the CNN model depends on the specified window size $k$ for the feature map $\mathbf{f}$, we need to pad the input matrix $X$ with $\lfloor \frac{k}{2} \rfloor$ zero column vectors on both sides to ensure the same fixed length $n$ for all the hidden vectors: $\mathbf{s}^\mathbf{f} = [s^\mathbf{f}_1, s^\mathbf{f}_2, \ldots, s^\mathbf{f}_n]$. Besides, we need to re-arrange the scores in the hidden vectors from different feature maps of the CNN so they are grouped according to the positions in the sentence, thus being compatible with the input requirement of RNNs. 

%In particular, let $F = \{\mathbf{f}_1, \mathbf{f}_2, \ldots, \mathbf{f}_m\}$ be the set of all feature maps for all the window sizes being used in the CNN model ($m = | F |$), the re-arranged hidden vector $h_i$ at position $i$ to be used by RNNs is computed by: $h_i = [s^{\mathbf{f}_1}_i, s^{\mathbf{f}_2}_i, \ldots, s^{\mathbf{f}_m}_i]$.

\subsubsection{Voting}

\label{sec:voting}

Instead of integrating CNNs and RNNs at the model level as the two previous methods, the voting method makes decision for a relation mention $X$ by voting the individual decisions of the different models. While there are several voting schemes in the literature, for this work, we employ the simplest scheme of majority voting. If there are more than one relation classes receiving the highest number of votes, the relation class returned by a model and having the highest probability would be chosen.

\subsection{The Hybrid Models}

\label{sec:hybrid}

%Gormley et al. \shortcite{Gormley:15} propose to further improve the performance of the neural networks for RE by integrating the traditional log-linear model that relies on various linguistic features from the past research on RE \cite{Zhou:05,Sun:11} 
%A potential way to further improve the performance of the neural networks for RE is to integrate the traditional log-linear model that relies on various linguistic features from the past research on RE \cite{Zhou:05,Sun:11} as do for the FCM model.

%probabilities for the relation classes obtained by the neural network models are multiplied by the corresponding scores in the traditional log-linear model 

In order to further improve the RE performance of models above, we investigate the integration of these neural network models with the traditional log-linear model that relies on various linguistic features from the past research on RE \cite{Zhou:05,Sun:11,Gormley:15}. Specifically, in such integration models (called {\it the hybrid models}), the relation class distribution is obtained from the element-wise multiplication between the distributions of the neural network models and the log-linear model. Let us take the ensembling model in Section \ref{sec:ensembling} as an example. The corresponding hybrid model in this case would be:
$p_{\text{hybrid}}(y|X) = \frac{1}{Z}p_{\text{C}}(y|X) p_{\text{R}}(y|X)p_{\text{login}}(y|X)$, assuming $p_{\text{login}}(y|X)$ be the distribution of the log-linear model and $Z$ be the normalization constant. The parameters of the log-linear model are learnt jointly with the parameters of the neural networks.

%(like Gormley et al. \shortcite{Gormley:15} do)

%A special note is that in the voting models of Section \ref{sec:voting}, we first integrate the separate CNN and RNN models with the log-linear model before we implement the voting procedure on the resulting models.

%The distinction between the neural network models above and the traditional log-linear model is that the former focuses more on the abstract meaning embedded in the relation mentions via aggregating embeddings of the individual words. The latter, on the other hand, relies on various linguistic properties mostly represented in the symbolic forms such as syntax or lexicon. Consequently, in this work, we hypothesize that the log-linear model would capture different information from the neural network models, causing the hybrid models to 

{\it Hypothesis}: Let $S$ be the set of relation mentions correctly predicted by some neural network model in some dataset (the coverage set). The introduction of the log-linear model into this neural network model essentially changes the coverage set of the network, resulting in the new coverage set $S'$ that might or might not subsume the original set $S$. In this work, we hypothesize that although $S$ and $S'$ overlap, there are still some relation mentions that only belong to either set. Consequently, we propose to implement a majority voting system (called the {\it hybrid-voting system}) on the outputs of the network and its corresponding hybrid model to enhance both models. 

Note that the voting models in Section \ref{sec:voting} involve the voting on two models (i.e, CNN and RNN). In order to integrate the log-linear model into such voting models, we first augment the separate CNN and RNN models with the log-linear model before we perform the voting procedure on the resulting models. Finally, the corresponding {\it hybrid-voting} systems would involve the voting on four models (CNN, hybrid CNN, RNN and hybrid RNN).

%Note that the voting models of Section \ref{sec:voting} already involve the voting on two models (i.e, CNN and RNN) and their corresponding hy

%report the performance of the systems that execute a majority voting on the outputs of a neural network model and its corresponding hybrid model. Note that the VOTE-BIDIRECT (or VOTE-BACKWARD) model as well as its corresponding hybrid model involve a voting on two models. Consequently, the voting systems of the VOTE-BIDIRECT (or VOTE-BACKWARD) model and its corresponding hybrid model would involve the voting on four models.

\subsection{Training}

We train the models by minimizing the negative log-likelihood function using the stochastic gradient descent algorithm with shuffled mini-batches and the AdaDelta update rule \cite{Zeiler:12,Kim:14}. The gradients are computed via back-propagation while regularization is executed by a dropout on the hidden vectors before the the multilayer neural networks \cite{Hinton:12}. During training, besides the weight matrices, we also optimized the embedding tables  $E, D, T, Q$ to achieve the optimal state. Finally, we rescale the weights whose $l_2$-norms exceed a hyperparameter \cite{Kim:14,Nguyen:15a}. 

\section{Experiments}

\subsection{Resources and Parameters}

%\footnote{\url{https://code.google.com/p/word2vec}}

For all the experiments below, we utilize the pre-trained word embeddings $\texttt{word2vec}$ with 300 dimensions from Mikolov et al. \shortcite{Mikolov:13} to initialize the word embedding table $E$. The parameters for CNNs and traning the networks are inherited from the previous studies, i.e, the window size set for feature maps  = $\{2,3,4,5\}$, 150 feature maps for each window size, 50 dimensions for all the embedding tables (except the word embedding table $E$), the dropout rate $ = 0.5$, the mini-batch size $ = 50$, the hyperparameter for the $l_2$ norms = 3 \cite{Kim:14,Nguyen:15a}. Regarding RNNs, we employ 300 units in the hidden layers.

\subsection{Dataset}

We evaluate our models on two datasets: the ACE 2005 dataset for relation extraction and the SemEval-2010 Task 8 dataset \cite{Hendrickx:10} for relation classification.

The ACE 2005 corpus comes with 6 different domains: broadcast conversation ($\texttt{bc}$), broadcast news ($\texttt{bn}$), telephone conversation ($\texttt{cts}$), newswire ($\texttt{nw}$),  usenet ($\texttt{un}$) and webblogs ($\texttt{wl}$). Following the common practice of domain adaptation research on this dataset \cite{Plank:13,Nguyen:14,Nguyen:15c,Gormley:15}, we use $\texttt{news}$ (the union of $\texttt{bn}$ and $\texttt{nw}$) as the training data, a half of $\texttt{bc}$ as the development set and the remainder ($\texttt{cts}$, $\texttt{wl}$ and the other half of $\texttt{bc}$) as the test data. Note that we are using the data prepared by Gormley et. al \shortcite{Gormley:15}, thus utilizing the same data split on $\texttt{bc}$ as well as the same data processing and NLP toolkits. The total number of relations in the training set is 43,497\footnote{It was an error in Gormley et al. \shortcite{Gormley:15} that reported 43,518 total relations in the training set. The authors acknowledged this error.}. We employ the BIO annotation scheme to capture the chunking information for words in the sentences and only mark the entity types of the two entity mention heads (obtained from human annotation) for this dataset.

The SemEval dataset concerns the relation classification task that aims to determine the relation type (or no relation) between two entities in sentences. In order to make it compatible with the previous research \cite{Socher:12,Gormley:15}, for this dataset, besides the word embeddings and the distance embeddings, we apply the name tagging, part of speech tagging and WordNet features (inherited from Socher et al. \shortcite{Socher:12} and encoded by the real-valued vectors for each word). The other settings are also adopted from the past studies \cite{Socher:12,Xu:15}.

%The SemEval dataset concerns the relation classification task that aims to determine the relation type (or no relation) between two entities in sentence. In order to make it compatible with the previous research \cite{Socher:12,Gormley:15}, for this dataset, instead of using the chunking features $q_i$, the golden entity type features $t_i$ and the features from dependency trees $p_i, g_i$, we apply the name tagging, part of speech tagging and WordNet features (inherited from Socher et al. \shortcite{Socher:12} and encoded by the real-valued vectors for each word in the sentences as the vectors $q_i$ and $t_i$). The other settings are also adopted from the past studies \cite{Socher:12,Xu:15}.

%Finally, we apply the feature set presented in the past RE research for the log-linear model \cite{Zhou:05,Sun:11,Gormley:15}.

\subsection{RNN Architectures}

This section evaluates the performance of various RNN architectures for RE on the development set. In particular, we compare different design combinations of the four following factors: (i) sentence representations (i.e, SEQ or DEP), (ii) transformation functions $\Phi$ (i.e, FF or GRU), (iii) the strategies to employ the hidden vector sequence for RE (i.e, HEAD or MAX), and (iv) the directions to run RNNs (i.e, forward ($\rightarrow$), backward ($\leftarrow$) or bidirectional ($\rightleftharpoons$)). Table \ref{tab:rnn} presents the results.

\begin{table}[htbp]
\small
\centering
\resizebox{.32\textwidth}{!}{
\begin{tabular}{|l|l|c|c|c|}
\hline
 \multicolumn{3}{|c|}{Systems} & DEP & SEQ \\ \hline
\multicolumn{1}{|c|}{} & \multicolumn{1}{c|}{} & $\rightleftharpoons$ & 60.78 & 63.22 \\ \cline{3-5}
\multicolumn{ 1}{|l|}{} & \multicolumn{ 1}{l|}{HEAD} & $\rightarrow$ & 55.55 & 60.05 \\ \cline{ 3- 5}
\multicolumn{ 1}{|l|}{FF} & \multicolumn{ 1}{l|}{} & $\leftarrow$ & 57.69 & 58.54 \\ \cline{ 2- 5}
\multicolumn{ 1}{|l|}{} & \multicolumn{ 1}{c|}{} & $\rightleftharpoons$ & 50.00 & 51.22 \\ \cline{ 3- 5}
\multicolumn{ 1}{|l|}{} & \multicolumn{ 1}{l|}{MAX} & $\rightarrow$ & 52.08 & 53.96 \\ \cline{ 3- 5}
\multicolumn{ 1}{|l|}{} & \multicolumn{ 1}{l|}{} & $\leftarrow$ & 45.07 & 33.50 \\ \cline{ 1- 5}
\multicolumn{ 1}{|c|}{} & \multicolumn{ 1}{c|}{} & $\rightleftharpoons$ & 63.32 & 63.23 \\ \cline{ 3- 5}
\multicolumn{ 1}{|l|}{} & \multicolumn{ 1}{l|}{HEAD} & $\rightarrow$ & 63.69 & 62.77 \\ \cline{ 3- 5}
\multicolumn{ 1}{|l|}{GRU} & \multicolumn{ 1}{l|}{} & $\leftarrow$ & 61.57 & 62.55 \\ \cline{ 2- 5}
\multicolumn{ 1}{|l|}{} & \multicolumn{ 1}{c|}{} & $\rightleftharpoons$ & 60.96 & {\bf 64.24} \\ \cline{ 3- 5}
\multicolumn{ 1}{|l|}{} & \multicolumn{ 1}{l|}{MAX} & $\rightarrow$ & 61.97 & {\bf 64.59} \\ \cline{ 3- 5}
\multicolumn{ 1}{|l|}{} & \multicolumn{ 1}{l|}{} & $\leftarrow$ & 61.56 & {\bf 64.30} \\ \hline
\end{tabular}
}
\caption{Performance (F1 scores) of RNNs on the dev set}
\label{tab:rnn}
\end{table}

\begin{comment}
\begin{table}[htbp]
\small
\centering
\begin{tabular}{|c|c|c|c|}
\hline
\multicolumn{1}{|l|}{} & \multicolumn{1}{l|}{} & SEQ & DEP \\ \hline
ff & head & 63.22 & 60.78 \\ \hline
\multicolumn{ 1}{|c|}{} & \multicolumn{ 1}{c|}{} & 60.05 & 55.55 \\ \cline{ 3- 4}
\multicolumn{ 1}{|c|}{} & \multicolumn{ 1}{c|}{} & 58.54 & 57.69 \\ \cline{ 2- 4}
\multicolumn{ 1}{|c|}{} & \multicolumn{ 1}{c|}{} & 51.22 & 50 \\ \cline{ 3- 4}
\multicolumn{ 1}{|c|}{} & \multicolumn{ 1}{c|}{} & 53.96 & 52.08 \\ \cline{ 3- 4}
\multicolumn{ 1}{|c|}{} & \multicolumn{ 1}{c|}{} & 33.5 & 45.07 \\ \hline
\multicolumn{ 1}{|c|}{} & \multicolumn{ 1}{c|}{} & 63.23 & 63.32 \\ \cline{ 3- 4}
\multicolumn{ 1}{|c|}{} & \multicolumn{ 1}{c|}{} & 62.77 & 63.69 \\ \cline{ 3- 4}
\multicolumn{ 1}{|c|}{} & \multicolumn{ 1}{c|}{} & 62.55 & 61.57 \\ \cline{ 2- 4}
\multicolumn{ 1}{|c|}{} & \multicolumn{ 1}{c|}{} & 64.24 & 60.96 \\ \cline{ 3- 4}
\multicolumn{ 1}{|c|}{} & \multicolumn{ 1}{c|}{} & 64.59 & 61.97 \\ \cline{ 3- 4}
\multicolumn{ 1}{|c|}{} & \multicolumn{ 1}{c|}{} & 64.3 & 61.56 \\ \hline
\end{tabular}
\caption{Performance of RNNs on the dev set}
\label{tab:rnn}
\end{table}
\end{comment}

The main conclusions include:

(i) Assuming the same choices for the other three corresponding factors, GRU is more effective than FF, SEQ is better than DEP most of the time and HEAD outperforms MAX (except the case where SEQ and GRU are applied) for RE with RNNs.

(ii) Regarding the direction mechanisms,  the bidirectional mechanism achieves the best performance for the HEAD strategy while the forward direction is the best mechanism for the MAX strategy. This can be partly explained by the lack of past or future context information in the HEAD strategy when we follow the backward or forward direction respectively.

%Table \ref{tab:rnn} presents the results. The main conclusions include:

%(i) Assuming the same conditions, the RE performance of the RNN models with GRU is consistently better than that with the FF transformation function, verifying the importance of the memory units (i.e, GRU) to avoid the ``{\it vanishing/exploding gradient}'' problem for RE with RNNs.

%(ii) The HEAD strategy outperforms the MAX strategy for RE given the same choices for the other three factors. The only exception for this statement is the RNN models applying the SEQ representation and the GRU function.

%(iii) Regarding the direction mechanisms, when the FF function is applied, the bidirectional mechanism achieves the best performance for the HEAD strategy while the forward direction is the best mechanism for the MAX strategy. This can be partly explained by the lack of past or future context information in the HEAD strategy when we follow the backward or forward direction respectively. Also, we can still observe such phenomenon when we employ the GRU function although the performance gap is less substantial.

%(iv) Finally, comparing the sentence representations SEQ and DEP, we see that SEQ has better performance than DEP most of the time, possibly stemming from the domination of relation mentions with short distance between entity mentions in the ACE 2005 dataset.

The best performance corresponds to the application of the SEQ representation, the GRU function and the MAX strategy that would be used in all the RNN models below. We call such RNN models with the forward, backward and bidirectional mechanism FORWARD, BACKWARD and BIDIRECT respectively. We also apply the SEQ representation for the CNN model (called CNN) in the following experiments for consistency.

\subsection{Evaluating the Combined Models}

\begin{table}[htbp]
\small
\centering
%\addtolength{\abovecaptionskip}{-1mm}
\addtolength{\belowcaptionskip}{-.5mm}
\resizebox{.36\textwidth}{!}{
\begin{tabular}{|l|c|c|c|}
\hline
\multicolumn{1}{|c|}{Model} & P & R & F1 \\ \hline
BIDIRECT & 69.16 & 59.97 & 64.24 \\ \hline
FORWARD & 69.33 & 60.45 & 64.59 \\ \hline
BACKWARD & 65.60 & 63.05 & 64.30 \\ \hline
CNN & 68.35 & 59.16 & 63.42 \\ \hline
%\rowcolor{Gray}
\multicolumn{4}{|l|}{Ensembling}   \\ \hline
%CNN+rnnHead & 69.54 & 61.43 & 65.23 \\ \hline
%CNN+rnnHeadForward & 69.2 & 54.62 & 61.05 \\ \hline
%CNN+rnnHeadBackward & 69.19 & 57.86 & 63.02 \\ \hline
CNN-BIDIRECT & 71.22 & 54.13 & 61.51 \\ \hline
CNN-FORWARD & 66.19 & 59.64 & 62.75 \\ \hline
CNN-BACKWARD & 65.09 & 60.13 & 62.51 \\ \hline
%\rowcolor{Gray}
\multicolumn{4}{|l|}{Stacking}  \\ \hline
%CNN+rnnHead & 62.78 & 67.26 & 64.95 \\ \hline
%CNN+rnnHeadForward & 66.48 & 57.54 & 61.69 \\ \hline
%CNN+rnnHeadBackward & 60.4 & 63.53 & 61.93 \\ \hline
CNN-BIDIRECT & 66.55 & 59.97 & 63.09 \\ \hline
CNN-FORWARD & 69.46 & 63.05 & {\bf 66.10} \\ \hline
CNN-BACKWARD & 72.58 & 58.35 & 64.69 \\ \hline
BIDIRECT-CNN & 65.63 & 61.59 & 63.55 \\ \hline
FORWARD-CNN & 73.13 & 58.67 & 65.11 \\ \hline
BACKWARD-CNN & 67.60 & 58.51 & 62.73 \\ \hline
%\rowcolor{Gray}
\multicolumn{4}{|l|}{Voting} \\ \hline
CNN-BIDIRECT & 71.08 & 60.94 & {\bf 65.62} \\ \hline
CNN-FORWARD & 70.38 & 59.32 & 64.38 \\ \hline
CNN-BACKWARD & 69.78 & 61.75 & {\bf 65.52} \\ \hline
\end{tabular}
}
\label{tab:combined}
\caption{Performance of the Combination Methods}
\end{table}

\begin{table*}[htbp]
\centering
%\addtolength{\abovecaptionskip}{-1mm}
%\addtolength{\belowcaptionskip}{-1mm}
\resizebox{0.73\textwidth}{!}{
\begin{tabular}{|l|c|c|g|c|c|g|c|c|g|}
\hline
\multicolumn{1}{|c|}{Model} & \multicolumn{ 3}{c|}{Neural Networks} & \multicolumn{ 3}{c|}{Hybrid Models} & \multicolumn{ 3}{c|}{Hybrid-Voting Models}   \\ \cline{2-10}
 & P & R & F1 & P & R & F1 & P & R & F1 \\ \hline
 CNN & 68.35 & 59.16 & 63.42 & 66.44 & 64.51 & 65.46 & 69.07 & 63.70 & 66.27 \\ \hline
BIDIRECT & 69.16 & 59.97 & 64.24 & 68.04 & 59.00 & 63.19 & 71.13 & 60.29 & 65.26 \\ \hline
FORWARD & 69.33 & 60.45 & 64.59 & 66.11 & 63.86 & 64.96 & 72.69 & 61.26 & 66.49 \\ \hline
BACKWARD & 65.60 & 63.05 & 64.30 & 66.03 & 62.07 & 63.99 & 71.56 & 63.21 & 67.13 \\ \hline
\multicolumn{10}{|l|}{Combined Models} \\ \hline
VOTE-BIDIRECT & 71.08 & 60.94 & 65.62 & 69.24 & 62.40 & 65.64 & 71.30 & 62.40 & {\bf 66.55} \\ \hline
STACK-FORWARD & 69.46 & 63.05 & 66.10 & 65.93 & 68.07 & 66.99 & 69.32 & 66.29 & {\bf 67.77} \\ \hline
VOTE-BACKWARD & 69.78 & 61.75 & 65.52 & 67.30 & 63.05 & 65.10 & 70.79 & 64.02 & {\bf 67.23} \\ \hline
\end{tabular}
}
\caption{Performance of the Hybrid Models}
\label{tab:hybrid}
\end{table*}

%combination

We evaluate the combination methods for CNNs and RNNs presented in Section \ref{sec:combinedModels}. In particular, for each method, we examine three models that are combined from one of the three RNN models FORWARD, BACKWARD, BIDIRECT and the CNN model. For instance, in the stacking method, the three combined models corresponding to the {\it RNN-CNN}  variant are FORWARD-CNN, BACKWARD-CNN, BIDIRECT-CNN while the three combined models corresponding to the {\it CNN-RNN} variant are CNN-FORWARD, CNN-BACKWARD, CNN-BIDIRECT. The notations for the other methods are self-explained. The model performance on the development set is given in Table \ref{tab:combined} that also includes the performance of the separate models (i.e, CNN, FORWARD, BACKWARD, BIDIRECT) for convenient comparison.

The first observation is that the ensembling method is not an effective way to combine CNNs and RNNs as its performance is worse than the separate models. Second, regarding the stacking method, the best way to combine CNNs and RNNs in this framework is to assemble the CNN model and the FORWARD model. In fact, the combination of the CNN and FORWARD models helps to improve the performance of the separate models in both variants of this method (referring to the models CNN-FORWARD and FORWARD-CNN). Finally, the voting method is also helpful as it outperforms the separate models with the CNN-BIDIRECT and CNN-BACKWARD combinations. 

For the following experiments, we would only focus on the three best combined models in this section, i.e, the CNN-FORWARD model in the stacking method (called STACK-FORWARD) and the CNN-BIDIRECT, CNN-BACKWARD models in the voting methods (called VOTE-BIDIRECT and VOTE-BACKWARD respectively).

\begin{table*}[htbp]
\centering
%\addtolength{\abovecaptionskip}{-1mm}
\addtolength{\belowcaptionskip}{-3mm}
\resizebox{0.82\textwidth}{!}{
\begin{tabular}{|l|c|c|c|c|c|c|c|c|c|c|}
\hline
 System & \multicolumn{ 3}{c|}{bc} & \multicolumn{ 3}{c|}{cts} & \multicolumn{ 3}{c|}{wl} &  \\ \cline{ 2- 11}
 & P & R & F1 & P & R & F1 & P & R & F1& Ave \\ \hline
FCM & 66.56 & 57.86 & 61.9 & 65.62 & 44.35 & 52.93 & 57.80 & 44.62 & 50.36 & 55.06 \\ \hline
Hybrid FCM & 74.39 & 55.35 & 63.48 & 74.53 & 45.01 & 56.12 & 65.63 & 47.59 & 55.17 & 58.26 \\ \hline \hline
%\rowcolor{Gray}
\multicolumn{11}{|l|}{Separate Systems}  \\ \hline
%Convolutional & \multicolumn{1}{l|}{} & \multicolumn{1}{l|}{} & \multicolumn{1}{l|}{} &  &  &  &  &  &  &  \\ \hline
Log-Linear & 68.44 & 50.07 & 57.83 & 73.62 & 41.57 & 53.14 & 60.40 & 47.31 & 53.06 & 54.68 \\ \hline
CNN & 65.62 & 61.06 & 63.26 & 65.92 & 48.12 & 55.63 & 54.14 & 53.68 & 53.91 & 57.60 \\ \hline
%%Hybrid CNN & 64.07 & 65.23 & 64.65 & 62.12 & 53.68 & 57.59 & 55.59 & 56.37 & 55.98 & 59.41 \\ \hline \hline
%RNN & \multicolumn{1}{l|}{} & \multicolumn{1}{l|}{} & \multicolumn{1}{l|}{} &  &  &  &  &  &  &  \\ \hline
BIDIRECT & 65.23 & 61.06 & 63.07 & 66.15 & 49.26 & 56.47 & 55.91 & 51.56 & 53.65 & 57.73 \\ \hline
%%Hybrid BIDIRECT & 65.77 & 60.92 & 63.25 & 67.4 & 50.08 & 57.46 & 57.10 & 53.54 & 55.26 & 58.66 \\ \hline \hline
FORWARD & 63.64 & 59.39 & 61.44 & 60.12 & 50.57 & 54.93 & 55.54 & 54.67 & 55.10 & 57.16 \\ \hline
%%Hybrid FORWARD & 62.52 & 62.87 & 62.69 & 63.15 & 47.95 & 54.51 & 51.74 & 52.55 & 52.14 & 56.45 \\ \hline \hline
BACKWARD & 60.44 & 61.2 & 60.82 & 58.20 & 54.01 & 56.03 & 51.03 & 52.55 & 51.78 & 56.21 \\ \hline
%%Hybrid BACKWARD & 62.76 & 59.53 & 61.10 & 65.45 & 49.92 & 56.64 & 55.27 & 54.96 & 55.11 & 57.62 \\ \hline \hline
%\small VOTE-BIDIRECT & 67.75 & 61.06 & 64.23 & 66.44 & 47.95 & 55.7 & 59.3 & 52.83 & 55.88 & 58.6 \\ \hline
%\small Hybrid VOTE-BIDIRECT & 65.64 & 62.45 & 64.01 & 65.82 & 51.06 & 57.51 & 58.23 & 54.11 & 56.09 & 59.2 \\ \hline
%\small STACK-FORWARD & 66.17 & 62.03 & 64.03 & 61.02 & 53.03 & 56.74 & 54.79 & 55.1 & 54.94 & 58.57 \\ \hline
%\small Hybrid STACK-FORWARD & 62.76 & 67.04 & 64.83 & 61.09 & 51.39 & 55.82 & 53.14 & 57.51 & 55.24 & 58.63 \\ \hline
%Combined & \multicolumn{1}{l|}{} & \multicolumn{1}{l|}{} & \multicolumn{1}{l|}{} &  &  &  &  &  &  &  \\ \hline
%\small VOTE-BACKWARD & 66.62 & 61.61 & 64.02 & 64.17 & 50.41 & 56.46 & 55.54 & 53.26 & 54.37 & 58.28 \\ \hline
%\small Hybrid VOTE-BACKWARD & 64.98 & 62.45 & 63.69 & 65.29 & 51.72 & 57.72 & 56.3 & 55.1 & 55.69 & 59.03 \\ \hline
%\rowcolor{Gray}
\multicolumn{11}{|l|}{Hybrid-Voting Systems}  \\ \hline
\small VOTE-BIDIRECT & 70.40 & 63.84 & {\bf 66.96\dag} & 66.74 & 49.92 & 57.12\dag & 59.24 & 54.96 & 57.02\dag & 60.37 \\ \hline
\small STACK-FORWARD & 65.75 & 66.48 & 66.11\dag & 63.58 & 51.72 & 57.04\dag & 56.35 & 57.22 & 56.78\dag & 59.98 \\ \hline
\small VOTE-BACKWARD & 69.57 & 63.28 & 66.28\dag & 65.91 & 52.21 & {\bf 58.26\dag} & 58.81 & 55.81 & {\bf 57.27\dag} & {\bf 60.60} \\ \hline
\end{tabular}
}
\caption{Comparison to the State-of-the-art. The cells marked with \dag designates the models that are significantly better than the other neural network models  ($\rho < 0.05$) on the corresponding domains.}
\label{tab:fcm}
\end{table*}

\subsection{Evaluating the Hybrid Models}

This section investigates the hybrid and {\it hybrid-voting} models (Section \ref{sec:hybrid}) to see if they can further improve the performance of the neural network models. In particular, we evaluate the separate models: CNN, BIDIRECT, FORWARD, BACKWARD, and the combined models: STACK-FORWARD, VOTE-BIDIRECT and VOTE-BACKWARD when they are augmented with the traditional log-linear model (the hybrid models). Besides, in order to verify the hypothesis in Section \ref{sec:hybrid}, we also test the corresponding {\it hybrid-voting} models. The experimental results are shown in Table \ref{tab:hybrid}. There are three main conclusions:

%The performance of the seven neural networks above is presented in column ``{\it Neural Networks}'' of Table \ref{tab:hybrid} while the corresponding hybrid models are shown in column ``{\it Hybrid Models}''. In column ``{\it Voted}'' of Table \ref{tab:hybrid}, we further report the performance of the systems that execute a majority voting on the outputs of a neural network model and its corresponding hybrid model. Note that the VOTE-BIDIRECT (or VOTE-BACKWARD) model as well as its corresponding hybrid model involve a voting on two models. Consequently, the voting systems of the VOTE-BIDIRECT (or VOTE-BACKWARD) model and its corresponding hybrid model would involve the voting on four models.

%There are three main conclusions from Table \ref{tab:hybrid}: 

(i) For all the models in columns ``{\it Neural Networks}'', ``{\it Hybrid Models}'' and ``{\it Hybrid-Voting Models}'', we see that the combined models outperform their corresponding separate models (only except the hybrid model of VOTE-BACKWARD), thereby further confirming the benefits of the combined models.

(ii) Comparing columns ``{\it Neural Networks}'' and ``{\it Hybrid Models}'', we find that the traditional log-linear model significantly helps the CNN model. The effects on the other models are not clear.

(iii) More interestingly, for all the neural networks being examined (either separate or combined), the corresponding {\it hybrid-voting} systems substantially improve both the neural network models as well as the corresponding hybrid models, testifying to the hypothesis about the {\it hybrid-voting} approach in Section \ref{sec:hybrid}. Note that the simpler voting systems on three models: the log-linear model, the CNN model and some RNN model (i.e, either BIDIRECT, FORWARD or BACKWARD) produce the worse performance than the {\it hybrid-voting} methods (the respective performance is 66.13\%, 65.27\%, and 65.96\%).

%The {\it hybrid-voting} models associated with the STACK-FORWARD, VOTE-BIDIRECT and VOTE-BACKWARD models are called HV-STACK-FORWARD, HV-VOTE-BIDIRECT and HV-VOTE-BACKWARD respectively.

\subsection{Comparing to the State-of-the-art}

The state-of-the-art system on the ACE 2005 for the unseen domains has been the feature-rich compositional embedding model (FCM) and the hybrid FCM model from Gormley et al. \shortcite{Gormley:15}. In this section, we compare the proposed {\it hybrid-voting} systems with these state-of-the-art systems on the test domains $\texttt{bc}$, $\texttt{cts}$, $\texttt{wl}$. Table \ref{tab:fcm} reports the results. For completeness, we also include the performance of the log-linear model and the separate models CNN, BIDIRECT, FORWARD, BACKWARD, serving as the other baselines for this work.

%and their corresponding hybrid systems

%Considering the hybrid models, the CNN model already outperforms the FCM model across all domains while the BIDIRECT model has a comparable performance and the FORWARD, BACKWARD models are slightly worse with FCM. Most importantly, the {\it hybrid-voting} systems are significantly better than the FCM models across all domains (up to 2\% improvement on the average absolute F score), {\it yielding the state-of-the-art performance for the unseen domains in this dataset}.
From the table, we see that although the separate neural networks outperform the FCM model across domains, they are still worse than the hybrid FCM model due to the introduction of the log-linear model into FCM. However, when the networks are combined and integrated with the log-linear model, they (the {\it hybrid-voting} systems) become significantly better than the FCM models across all domains (up to 2\% improvement on the average absolute F score), {\it yielding the state-of-the-art performance for the unseen domains in this dataset}.

\begin{comment}
\begin{table}[htbp]
\small
\centering
\resizebox{.5\textwidth}{!}{
\begin{tabular}{|l|c|c|c|c|c|c|}
\hline
 & bc &  & cts &  & wl &  \\ \hline
 & CNN & RNN & CNN & RNN & CNN & RNN \\ \hline
PHYS & 41.9 & 48.9 & 35.1 & 42.7 & 33.1 & 40.4 \\ \hline
PART-WHOLE & 60.8 & 62.8 & 53.7 & 52.4 & 51.6 & 52.4 \\ \hline
ART & 61.7 & 60.9 & 64.9 & 70.6 & 51.4 & 53.3 \\ \hline
ORG-AFF & 79 & 78.8 & 62.4 & 64.7 & 59.6 & 57.6 \\ \hline
PER-SOC & 74.7 & 72.7 & 69.6 & 70.4 & 58.8 & 60.5 \\ \hline
GEN-AFF & 53.1 & 50.8 & 38.7 & 28.3 & 51.1 & 53.7 \\ \hline
all & 63.3 & 63.1 & 55.6 & 56.5 & 53.9 & 53.6 \\ \hline
\end{tabular}
}
\label{}
\caption{}
\end{table}
\end{comment}

\subsection{Relation Classification Experiments}

We further evaluate the proposed systems for the relation classification task on the SemEval dataset. Table \ref{tab:rc} presents the performance of the seprate models, the proposed systems as well as the other representative systems on this task. The most important observation is that the {\it hybrid-voting} systems VOTE-BIDIRECT  and VOTE-BACKWARD achieve the state-of-the-art performance for this dataset, further highlighting their benefit for relation classification. The {\it hybrid-voting} STACK-FORWARD system performs less effectively in this case, possibly due to the small size of the SemEval dataset that is not sufficient to training such a deep model.

\begin{table}[htbp]
\small
%\addtolength{\abovecaptionskip}{-2mm}
%\addtolength{\belowcaptionskip}{-2mm}
\centering
\resizebox{.34\textwidth}{!}{
\begin{tabular}{p{5cm}p{0.5cm}}
\hline
 \multicolumn{1}{l}{Classifier} & \multicolumn{1}{c}{F} \\ \hline \hline
%SVM & POS, stemming, syntactic patterns & 60.1 \\ \hline
%SVM & word pair, words in between & 72.5 \\ \hline
%SVM & POS, WordNet, stemming, syntactic patterns & 74.8 \\ \hline
%SVM & POS, WordNet, morphological features, thesauri, Google $n$-grams & 77.6 \\ \hline
%MaxEnt & POS, WordNet, morphological features, noun compound system, thesauri, Google $n$-grams & 77.6 \\ \hline
SVM \cite{Hendrickx:10} & 82.2 \\ \hline
%RNN & word embeddings, syntactic parse & 74.8 \\
RNN \cite{Socher:12} & 77.6 \\ 
%MVRNN & word embeddings, syntactic parse & 79.1 \\
MVRNN \cite{Socher:12} & 82.4 \\ \hline
%O-CNN & - & 78.9 \\
CNN \cite{Zeng:14} & 82.7 \\ \hline
%CR-CNN (log-loss) & word embeddings & 82.7 \\
CR-CNN \cite{Santos:15} & {\bf 84.1\dag} \\ \hline
FCM \cite{Gormley:15} & 83.0 \\
Hybrid FCM \cite{Gormley:15}  & 83.4 \\ \hline
DepNN \cite{Liu:15} & 83.6 \\ \hline
%SDP-LSTM & word embeddings & 82.4 \\
SDP-LSTM \cite{Xu:15} & 83.7 \\ \hline
CNN & 83.5 \\
BIDIRECT & 81.8 \\
FORWARD & 81.9 \\
BACKWARD & 82.4 \\ \hline
VOTE-BIDIRECT &  {\bf 84.1} \\
STACK-FORWARD & 83.4 \\
VOTE-BACKWARD & {\bf 84.1} \\
\end{tabular}
}
\caption{Performance of Relation Classification Systems. The \dag refers to special treatment to the $\texttt{Other}$ class.}
\label{tab:rc}
\end{table}

\subsection{Analysis}

%This section analyzes the outputs of the systems to gain more insights into their operation.

%- {\bf CNNs vs RNNs}: 
In order to better understand the reason helping the combination of CNNs and RNNs outperform the individual networks, we evaluate the performance breakdown per relation for the CNN and BIDIRECT models. The results on the development set of the ACE 2005 dataset are provided in Tabel \ref{tab:breakdown}. 

\begin{table}[htbp]
\centering
%\addtolength{\abovecaptionskip}{-1mm}
\addtolength{\belowcaptionskip}{-2mm}
\resizebox{.43\textwidth}{!}{
\begin{tabular}{|l|c|c|c|c|c|c|}
\hline
Relation Class& \multicolumn{3}{c|}{CNN} & \multicolumn{3}{c|}{BIDIRECT} \\ \cline{2-7}
& P & R & F1 & P & F & F1 \\ \hline
PHYS & 66.7 & 34.7 & 45.7 & 57.4 & 50.9 & 54.0 \\ \hline
PART-WHOLE & 68.6 & 67.8 & 68.2 & 74.4 & 70.1 & 72.2 \\ \hline
ART & 64.2 & 51.2 & 57.0 & 68.6 & 41.7 & 51.9 \\ \hline
ORG-AFF & 70.2 & 83.0 & 76.0 & 79.3 & 76.1 & 77.7 \\ \hline
PER-SOC & 71.1 & 59.3 & 64.6 & 69.6 & 59.3 & 64.0 \\ \hline
GEN-AFF & 65.9 & 55.1 & 60.0 & 59.0 & 46.9 & 52.3 \\ \hline
all & 68.4 & 59.2 & 63.4 & 69.2 & 60.0 & 64.2 \\ \hline
\end{tabular}
}
\caption{The Performance Breakdown per Relation for CNN and BIDIRECT on the development set.}
\label{tab:breakdown}
\end{table}

One of the main insights is although CNN and BIDIRECT have the comparable overall performance, their recalls on individual relations are very diverged. In particular, the BIDIRECT has much better recall for the PHYS relation while the recalls of CNN are significantly better for the ART, ORG-AFF and GEN-AFF relations. A closer investigation reveals two facts: (i) the PHYS relation mentions that are only correctly predicted by BIDIRECT involve the long distances between two entity mentions, such as the PHYS relation between ``{\it Some}'' (a person entity) and ``{\it desert}'' (a location entity) in the following sentence: ``{\it Some of the 40,000 British troops are kicking up a lot of dust in the Iraqi desert making sure that nothing is left behind them that could hurt them.}'', and (ii) the ART, ORG-AFF, GEN-AFF relation mentions only correctly predicted by CNN contains the patterns between the two entity mentions that are short but meaningful enough to decide the relation classes, such as ``{\it The Iraqi unit in possession of those guns}'' (the ART relation between ``{\it unit}'' and ``{\it guns}''), or ``{\it the al Qaeda chief operations officer}'' (the ORG-AFF relation between ``{\it al Qaeda}'' and ``{\it officer}''). The failure of CNN on the PHYS relation mentions with long distances originates from its mechanism to model short and consecutive $k$-grams (up to length 5 in our case), causing the difficulty to capture the long and/or unconsecutive patterns. BIDIRECT, on the other hand, fails to predict the short (but expressive enough) patterns for ART, ORG-AFF, GEN-AFF because it involves the hidden vectors that only model the context words outside the short patterns, potentially introducing unnecessary and noisy information into the max-pooling scores for prediction. Eventually, the combination of RNNs and CNNs helps to compensate the drawbacks of each model.

\section{Related Work}

%Neural networks have opened a new direction to solve many NLP tasks recently, starting from the first application 

Starting from the invention of the distributed representations for words \cite{Bengio:03,Mnih:08,Collobert:08,Turian:10,Mikolov:13}, CNNs and RNNs have gained significant successes on various NLP tasks, including sequential labeling \cite{Collobert:11}, sentence modeling and classification \cite{Kalchbrenner:14,Kim:14}, paraphrase identification \cite{Yin:15}, event extraction \cite{Nguyen:15b,Chen:15} for CNNs and machine translation \cite{Cho:14,Bahdanau:15} for RNNs, to name a few.

% sentiment analysis \cite{Socher:13}, question answering \cite{Mohit:14},
%investigate recurrent NNs along the shortest dependency paths

For relation extraction/classification, most work on neural networks has focused on the relation classification task. In particular, Socher et al. \shortcite{Socher:12} and Ebrahimi and Dou \shortcite{Ebrahimi:15} study the recursive NNs that recur over the tree structures while Xu et al. \shortcite{Xu:15} and Zhang and Wang \shortcite{Zhang:15} investigate recurrent NNs. Regarding CNNs, Zeng et al. \shortcite{Zeng:14} examine CNNs via the sequential representation of sentences, dos Santos et al. \shortcite{Santos:15} explore a ranking loss function with data cleaning while Zeng et al. \shortcite{Zeng:15} propose dynamic pooling and multi-instance learning. For RE, Yu et al. \shortcite{Yu:15} and Gormley et al. \shortcite{Gormley:15} work on the feature-rich compositional embedding models. Finally, the only work that combines NN architectures is due to Liu et al. \shortcite{Liu:15} but it only focuses on the stacking of the recursive NNs and CNNs for relation classification.

\section{Conclusion}

We investigate different methods to combine CNNs, RNNs as well as the hybrid models to integrate the log-linear model into the NNs. The experimental results demonstrate that the simple majority voting between CNNs, RNNs and their corresponding hybrid models is the best combination method. We achieve the state-of-the-art performance for both relation extraction and relation classification. In the future, we plan to further evaluate the proposed methods on the other tasks such as event extraction and slot filling in the KBP evaluation.

% on the ACE 2005 and SemEval dataset

\section*{Acknowledgment}

We would like to thank Matthew Gormley and Mo Yu for providing the dataset. Thank you to Kyunghyun Cho and Yifan He for valuable suggestions.

\bibliography{naaclhlt2016}
\bibliographystyle{naaclhlt2016}

\end{document}